\documentclass[conference]{IEEEtran}

\makeatletter

\def\ps@IEEEtitlepagestyle{%
  \def\@evenfoot{}%
}

\usepackage{eso-pic}
\IEEEoverridecommandlockouts
\usepackage{cite}
\usepackage{amsmath,amssymb,amsfonts}
\usepackage{algorithmic}
\usepackage{graphicx}
\usepackage{textcomp}
\usepackage{xcolor}
\usepackage{booktabs}
\usepackage{tikz}
\usepackage{url}
\usepackage{pgfplots}
\pgfplotsset{compat=1.17}
\usetikzlibrary{positioning,arrows.meta,calc}
\def\BibTeX{{\rm B\kern-.05em{\sc i\kern-.025em b}\kern-.08em
    T\kern-.1667em\lower.7ex\hbox{E}\kern-.125emX}}

\usepackage{eso-pic}
\begin{document}
\title{\vspace*{1cm} TurnTurk-GPT: Text-Based Turn-Taking Prediction for Turkish
}

\author{\IEEEauthorblockN{Ahmet Tuğrul Bayrak}
\IEEEauthorblockA{\textit{Data Science and Innovation} \\
\textit{Ata Technology Platforms}\\
Istanbul, Turkey \\
tugrul.bayrak@atptech.com}
\and
\IEEEauthorblockN{Fatma Nur Korkmaz}
\IEEEauthorblockA{\textit{Data Science and Innovation} \\
\textit{Ata Technology Platforms}\\
Istanbul, Turkey \\
fatmanur.korkmaz@atptech.com}
\and
\IEEEauthorblockN{Nur Yağmur Tuncer}
\IEEEauthorblockA{\textit{Data Science and Innovation} \\
\textit{Ata Technology Platforms}\\
Istanbul, Turkey \\
yagmur.tuncer@atptech.com}
\and
\IEEEauthorblockN{Mustafa Sertaç Türkel}
\IEEEauthorblockA{\textit{Data Science and Innovation} \\
\textit{Ata Technology Platforms}\\
Istanbul, Turkey \\
sertac.turkel@atptech.com}
}

%\author{Anonymous Submission}

\maketitle
{\let\thefootnote\relax\footnotetext{Accepted at ISAS 2026. \copyright~2026 IEEE. Personal use of this material is permitted. Permission from IEEE must be obtained for all other uses, in any current or future media, including reprinting/republishing this material for advertising or promotional purposes, creating new collective works, for resale or redistribution to servers or lists, or reuse of any copyrighted component of this work in other works.}}

\begin{abstract}
This paper presents TurnTurk-GPT, a text-based turn-taking
predictor for Turkish obtained by fine-tuning a Turkish GPT-2 model with the TurnGPT
objective on synthetic and real dialogue, and evaluates it on real Turkish
conversation against rule baselines, a zero-shot Turkish language model and a
widely used multilingual end-of-turn detector. Evaluation uses word-level
decision points on unpunctuated, lower-cased text approximating the output of
streaming speech recognisers, recording-disjoint and speaker-disjoint folds, nested
threshold selection and paired cluster bootstrap tests. TurnTurk-GPT outperforms the multilingual detector under both recording-disjoint and speaker-disjoint evaluation. Its gain over a zero-shot Turkish language model is largest for speakers seen in training, while on unseen speakers it matches this strong baseline. Removing the suffixes of the turn-final word
reduces AP by up to $44\%$, against $2\%$ for a sentence-internal word,
locating the Turkish end-of-turn cue in verb-final morphology. Synthetic
dialogue alone yields the weakest trained model, and when a speaker pauses the
zero-shot and multilingual baselines remain close to chance. The model and the
evaluation protocol are publicly released.
\end{abstract}


\section{Introduction}

Conversation proceeds through an orderly exchange of speaking turns, with gaps and overlaps measured in a few hundred milliseconds \cite{sacks1974simplest}. A spoken dialogue system that responds too early interrupts the user, whereas one that responds too late causes unnatural delays. The system therefore requires an incrementally updated estimate of whether the current position is a plausible turn ending \cite{schlangen2011incremental,skantze2021review}.

Text-based predictors treat this as a token-level prediction problem inside a language model \cite{ekstedt2020turngpt}; audio-based predictors learn future voice activity from the waveform \cite{ekstedt2022vap,inoue2024realtime}. Research models and benchmarks have focused on English, with multilingual work covering English, Mandarin and Japanese \cite{inoue2024multilingual}; a recent multi-domain benchmark is English-only \cite{jiang2026turnbench}. Although deployed multilingual end-of-turn detectors list Turkish among their supported languages \cite{livekit2025turndetector}, their performance on real Turkish conversation has not been reported under a public evaluation protocol. Turkish is verb-final and agglutinative, and its turn-final cues may differ substantially from those of English.

Two Turkish resources for this task were introduced in previous work: Syn-TurnTurk, a large corpus of dialogues generated by a large language model \cite{bayrak2026synturnturk}, and Real-TurnTurk, a small multimodal corpus of real dyadic conversations \cite{bayrak2026realturnturk}. Real-TurnTurk was collected because openly available Turkish speech resources do not include spontaneous two-party conversation with speaker-separated audio and time-aligned transcripts: conversational corpora are either commercial or restricted to academic access on request, and open corpora consist of read or single-speaker recordings. Building on these two resources, TurnTurk-GPT, a TurnGPT-style turn-taking predictor on a Turkish backbone, is trained under four configurations combining synthetic and real data. It is evaluated under a protocol defined for Turkish turn-taking prediction, which relies on word-level decision points, average precision, recording-disjoint and speaker-disjoint folds, nested threshold selection and paired cluster bootstrap tests. Within this protocol, TurnTurk-GPT is compared with a zero-shot Turkish language model read in four ways and with a multilingual end-of-turn detector used as released, and the conditions under which it improves on them are identified. Further analyses specific to Turkish show that the end-of-turn cue is carried by the suffixes of the final word, that the choice of tokeniser has a substantial effect, that synthetic dialogue alone provides a weak training signal, and that the text-only baselines remain close to chance level when the speaker pauses.

In addition, the model is publicly available on Hugging Face as a baseline for Turkish turn-taking prediction.\footnote{https://huggingface.co/tugrulbayrak/TurnTurk-Gpt}

\section{Related Work}

\subsection{Turn-taking prediction}

Early systems determined turn ends with silence thresholds, which tend to interrupt speakers who pause and to respond late to speakers who finish quickly
\cite{skantze2021review}. Continuous prediction replaced fixed thresholds:
Roddy et al.\ \cite{roddy2018investigating} predict speech activity over a
future window and find acoustic and lexical features complementary. Voice
Activity Projection \cite{ekstedt2022vap} learns discretised future
voice-activity patterns in a self-supervised way; it has been run in real time
\cite{inoue2024realtime} and extended to English, Mandarin and Japanese, where
cross-lingual fine-tuning was found to transfer \cite{inoue2024multilingual}.
Evaluation remains fragmented across corpora and metrics, which recent
benchmark efforts address for English \cite{jiang2026turnbench}.

\subsection{Text-based models}

TurnGPT \cite{ekstedt2020turngpt} adds a turn-shift token to a pre-trained
transformer language model \cite{vaswani2017attention,radford2019gpt2} and
trains it with the causal language modelling loss, so the turn-shift
probability at each word depends only on preceding context and can be computed
incrementally as words arrive. This formulation is adopted here. Production voice
agents combine silence detection with a text-based end-of-turn classifier that
is queried on the recogniser's transcript when the user stops speaking
\cite{livekit2025turndetector}; such a classifier is evaluated here as a baseline.
Because a pre-trained language model already assigns probabilities to sentence
and document boundaries, it can also be read as a zero-shot predictor without
any turn-taking training; this is included as a baseline that isolates what
task-specific fine-tuning adds.

\subsection{Turkish}

Turkish is agglutinative and predominantly verb-final: tense, person, mood and
question marking are expressed by suffixes on the clause-final verb, so the
information that signals completion is concentrated at the end of the
utterance and inside a single word. Monolingual Turkish transformer language
models have only recently become available \cite{kesgin2024cosmosgpt}; one of
them serves here as the backbone and as the zero-shot baseline. Publicly
available turn-taking resources for Turkish consist of the two corpora
introduced in previous work and used in this study
\cite{bayrak2026synturnturk,bayrak2026realturnturk}.

\section{Data}

Table~\ref{tab:data} summarises the corpora. Syn-TurnTurk\footnote{\url{https://huggingface.co/datasets/tugrulbayrak/Syn-TurnTurk}} contains topic-conditioned two-party dialogues
generated by a large language model \cite{bayrak2026synturnturk}. Real-TurnTurk\footnote{\url{https://huggingface.co/datasets/tugrulbayrak/Real-TurnTurk}}
contains six dyadic video-call recordings with six distinct speakers, separate
audio channels, per-channel voice activity detection, backchannel annotation and
transcripts from the platform's live captioning \cite{bayrak2026realturnturk}. Only the text and the
turn structure are used.

\begin{table}[t]
\caption{Data Information}
\label{tab:data}
\centering
\small
\begin{tabular}{|l|r|r|}
\hline
 & Syn-TurnTurk & Real-TurnTurk \\
\hline
Dialogues / recordings & 1{,}625 & 6 \\
\hline
Turns                  & 14{,}185 & 578 \\
\hline
Words                  & 123{,}397 & 15{,}531 \\
\hline
Mean words per turn    & 8.7 & 26.9 \\
\hline
\end{tabular}
\end{table}

Real turns are derived from voice activity: consecutive segments of one speaker
are merged into a floor-holding turn, which backchannels do not split, and
transcript segments are assigned to turns by temporal overlap. $4.7\%$ of words
lie in segments straddling a turn boundary and $0.5\%$ conflict with the voice
activity annotation and are discarded \cite{bayrak2026realturnturk}. Of 930
turn-internal pauses, 633 could be attributed to a word, with a
median duration of $0.50$~s and 297 longer than $0.5$~s; the remainder fall
inside a transcript segment and cannot be localised. These positions are the most likely points of interruption for a deployed system and are used as hard negatives throughout. The difference in turn length between the corpora
(Table~\ref{tab:data}) indicates that they describe different phenomena:
synthetic turns are short, complete sentences, whereas real turns are long,
contain internal pauses, and are frequently resumed after a backchannel.

\section{Method}

\subsection{Model and objective}

\begin{figure}[t]
\centering
\begin{tikzpicture}[
  font=\scriptsize,
  tok/.style={draw, rounded corners=1.5pt, minimum height=4.2mm,
              inner xsep=2.5pt, inner ysep=1.5pt},
  ts/.style={tok, fill=black!8, font=\scriptsize\ttfamily},
  spk/.style={font=\tiny, text=black!60}
]
\node[ts]                       (t1) {<ts>};
\node[tok, right=1pt of t1]     (t2) {bu};
\node[tok, right=1pt of t2]     (t3) {iş};
\node[tok, right=1pt of t3]     (t4) {bitti};
\node[ts,  right=1pt of t4]     (t5) {<ts>};
\node[tok, right=1pt of t5]     (t6) {evet};
\node[tok, right=1pt of t6]     (t7) {anladım};
\node[ts,  right=1pt of t7]     (t8) {<ts>};
\foreach \n in {t1,t2,t3,t4,t5}{\node[spk, below=1.2mm of \n] {sp1};}
\foreach \n in {t6,t7,t8}{\node[spk, below=1.2mm of \n] {sp2};}
\node[above=4.5mm of t3, font=\tiny] (s3) {$0.02$};
\node[above=4.5mm of t4, font=\tiny] (s4) {$\mathbf{0.61}$};
\node[above=4.5mm of t7, font=\tiny] (s7) {$\mathbf{0.74}$};
\draw[-{Latex[length=1.2mm]}, black!55] (t3.north) -- (s3.south);
\draw[-{Latex[length=1.2mm]}, black!55] (t4.north) -- (s4.south);
\draw[-{Latex[length=1.2mm]}, black!55] (t7.north) -- (s7.south);
\draw[black!45] ($(t2.south west)+(0,-4.2mm)$) -- ($(t4.south east)+(0,-4.2mm)$)
      node[midway, below, font=\tiny, text=black!60] {turn 1 (speaker A)};
\draw[black!45] ($(t6.south west)+(0,-4.2mm)$) -- ($(t7.south east)+(0,-4.2mm)$)
      node[midway, below, font=\tiny, text=black!60] {turn 2 (speaker B)};
\end{tikzpicture}
\caption{Serialisation and scoring}
\label{fig:serial}
\end{figure}
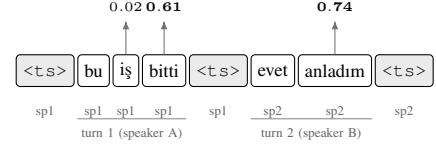

Following TurnGPT \cite{ekstedt2020turngpt}, a turn-shift token \texttt{<ts>}
and two speaker tokens are added to the vocabulary of
turkish-gpt2 (124M parameters) \cite{kesgin2024cosmosgpt}. A dialogue
is serialised as its turns, each closed by \texttt{<ts>}, with a leading
\texttt{<ts>}; speaker identity is supplied through \texttt{token\_type\_ids}.
Training uses the causal language modelling loss, and the score of a word is
$P(\texttt{<ts>})$ at its final subword token (Fig.~\ref{fig:serial}).

\subsection{Training configurations and input}\label{sec:input}

\textbf{S}: synthetic data only; \textbf{R}: real data only; \textbf{SR}: one
stage on both; \textbf{S2R}: synthetic pre-training, then real fine-tuning.
Optimisation uses a fixed step budget, since the corpora
differ in size by two orders of magnitude. Each configuration uses three seeds,
averaged per position.

The optimiser is AdamW with weight decay $0.01$, a linear schedule with $10\%$
warm-up, gradient clipping at $1.0$ and mixed precision. Training windows of 256
tokens overlap by half; at evaluation, every position is scored exactly once
with at least 32 tokens of left context. Embeddings of the added tokens are
initialised at the mean of the existing embeddings. The development set for
early stopping is the final $12\%$ of the windows of each training dialogue,
with the adjacent window discarded to prevent overlap; validation loss is
checked every 20 steps (50 for stages involving synthetic data) with a patience
of five checks. The synthetic stage holds out $5\%$ of the dialogues for this
purpose.

The input is lower-cased text with punctuation removed, which approximates the output of streaming recognisers at decision time and corresponds to the normalisation used by the multilingual detector described below. The transcripts are final rather than streaming hypotheses, so all results are offline.

\section{Evaluation}

\subsection{Protocol}

Every word of the real corpus is a decision point, labelled positive if it ends
its turn: $15{,}522$ points, $578$ positive. The primary metric is
average precision (AP) \cite{davis2006relationship}; a random predictor scores
$0.037$. Precision, recall and F1 are taken at thresholds selected on the other
folds (nested). Two splits are used: \emph{recording-disjoint}
(leave-one-recording-out), and \emph{speaker-disjoint}, in which no speaker of the
test recording appears in training (training shrinks to two or three
recordings). Confidence intervals are recording-level cluster bootstrap
intervals, and comparisons use a paired cluster bootstrap on the same decision
points \cite{efron1993bootstrap}; a difference is significant when the $95\%$
interval excludes zero. Each of 2{,}000 bootstrap replicates draws six
recordings with replacement and includes all decision points of every drawn
recording; intervals are the $2.5$th and $97.5$th percentiles of the replicate
statistics. In paired comparisons both systems are evaluated on the same
replicates. Because resampling is performed over recordings, the intervals
reflect variation between conversations rather than between individual decision points. The false alarm rate is also reported on the 633
\emph{pre-pause} negatives, words followed by a turn-internal pause.

\subsection{Baselines}

B0 is random; B1 is a turn-length prior estimated on the
training folds.

\textbf{Zero-shot language model:} The backbone without fine-tuning reads the
same normalised words, with turns separated by a newline. At each word, four
readings are taken:
$P(\text{next}\in\{.,?,!\})$, $P(\text{end of text})$, $P(\text{newline})$ and
their sum; the newline reading directly estimates the probability that the line, and therefore the turn, ends.

\textbf{Multilingual detector:} The LiveKit multilingual turn detector
\cite{livekit2025turndetector} is used as released, through the inference code
of its own framework, with its own normalisation (lower-casing, punctuation removal) and
context. Its input contains the current turn up to the word being scored.

\section{Results}

\subsection{Main comparison}

\begin{table}[t]
\caption{Main results}
\label{tab:main}
\centering
\small
\setlength{\tabcolsep}{3.5pt}
\begin{tabular}{|l|c|c|c|c|c|}
\hline
\textbf{System} & \textbf{AP} & \textbf{95\% CI} & \textbf{AUC} & \textbf{F1} & \textbf{PP} \\
\hline
B0 random            & .037 & [.028,.051] & .500 & .072 & 100\% \\
\hline
B1 length prior      & .049 & [.036,.064] & .572 & .089 & 12.3\% \\
\hline
LM $P(.?!)$          & .124 & [.100,.147] & .762 & .163 & 12.6\% \\
\hline
LM $P(\text{eot})$   & .076 & [.064,.091] & .676 & .132 & 18.8\% \\
\hline
LM $P(\text{newline})$ & .130 & [.103,.154] & .787 & .195 & 25.0\% \\
\hline
LiveKit detector     & .100 & [.081,.118] & .686 & .132 & 14.4\% \\
\hline
\hline
S \ \ synthetic      & .082 & [.068,.092] & .712 & .132 & 33.5\% \\
\hline
R \ \ real           & .116 & [.082,.159] & .738 & .163 & 18.3\% \\
\hline
SR mixed             & .120 & [.107,.126] & .767 & .189 & 17.4\% \\
\hline
S2R         & .148 & [.132,.167] & .783 & .219 & 13.1\% \\
\hline
\end{tabular}
\end{table}

\begin{table}[t]
\caption{Paired cluster bootstrap comparisons}
\label{tab:paired}
\centering
\small
\setlength{\tabcolsep}{4pt}
\begin{tabular}{|l|c|c|c|}
\hline
\textbf{Comparison} & \textbf{$\Delta$AP} & \textbf{95\% CI} & \textbf{Sig.} \\
\hline
S2R $-$ LiveKit             & $+.048$ & $[+.034,+.060]$ & yes \\
\hline
S2R $-$ LM newline          & $+.019$ & $[+.002,+.037]$ & yes \\
\hline
LiveKit $-$ LM newline      & $-.029$ & $[-.039,-.016]$ & yes \\
\hline
\multicolumn{4}{|l|}{\emph{Speaker-disjoint S2R}} \\
\hline
S2R $-$ LiveKit             & $+.020$ & $[+.009,+.034]$ & yes \\
\hline
S2R $-$ LM newline          & $-.010$ & $[-.025,+.011]$ & no \\
\hline
speaker-disj. $-$ rec.-disj. & $-.028$ & $[-.036,-.021]$ & yes \\
\hline
\multicolumn{4}{|l|}{\emph{Training configurations}} \\
\hline
S2R $-$ R                   & $+.032$ & $[-.008,+.063]$ & no \\
\hline
S2R $-$ SR                  & $+.028$ & $[+.013,+.044]$ & yes \\
\hline
SR $-$ R                    & $+.004$ & $[-.038,+.037]$ & no \\
\hline
\end{tabular}
\end{table}

\begin{figure}[t]
\centering
\begin{tikzpicture}
\begin{axis}[
  width=0.86\columnwidth, height=5.6cm,
  xbar, bar width=5pt, bar shift=0pt, xmin=0, xmax=0.18,
  ymin=-0.7, ymax=9.7,
  ytick={0,...,9}, yticklabels={{B1 length prior},{LM $P$(eot)},{S synthetic},{LiveKit detector},{R real},{SR mixed},{S2R, unseen speakers},{LM $P$(.?!)},{LM $P$(newline)},{S2R}},
  y tick label style={font=\scriptsize},
  x tick label style={font=\scriptsize, /pgf/number format/fixed, /pgf/number format/precision=2},
  xlabel={\scriptsize Average precision (95\% CI)},
  axis x line*=bottom, axis y line*=left,
  legend style={font=\tiny, at={(0.98,0.03)}, anchor=south east, draw=none},
  error bars/x dir=both, error bars/x explicit,
]
\addplot[fill=black!25, draw=black!60, error bars/error bar style={black!70}] coordinates {
      (0.049,0) += (0.015,0) -= (0.013,0)
      (0.076,1) += (0.015,0) -= (0.012,0)
      (0.100,3) += (0.018,0) -= (0.019,0)
      (0.124,7) += (0.023,0) -= (0.024,0)
      (0.130,8) += (0.024,0) -= (0.027,0)};
\addlegendentry{baselines}
\addplot[fill=black!70, draw=black, error bars/error bar style={black}] coordinates {
      (0.082,2) += (0.010,0) -= (0.014,0)
      (0.116,4) += (0.043,0) -= (0.034,0)
      (0.120,5) += (0.006,0) -= (0.013,0)
      (0.120,6) += (0.018,0) -= (0.015,0)
      (0.148,9) += (0.019,0) -= (0.016,0)};
\addlegendentry{TurnTurk-GPT configurations}
\draw[dashed, black!60] (axis cs:0.037,-0.7) -- (axis cs:0.037,9.7);
\end{axis}
\end{tikzpicture}
\caption{Average precision}
\label{fig:ap}
\end{figure}
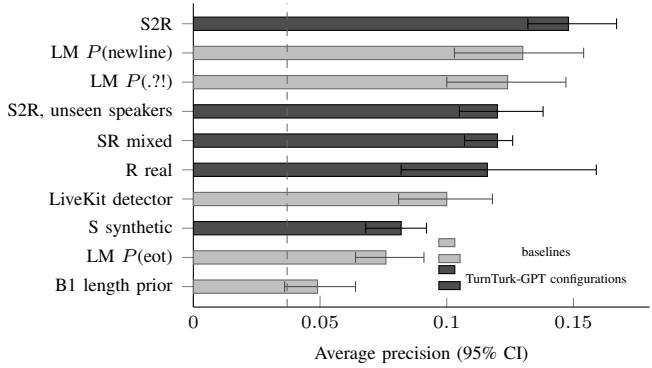

Table~\ref{tab:main} gives the results and
Table~\ref{tab:paired} the paired comparisons. Under the speaker-disjoint split,
S2R reaches AP $0.120$. Fig.~\ref{fig:ap} shows the same comparison graphically with recording-level 95\% bootstrap intervals; all bars refer to the recording-disjoint split except the one marked for unseen speakers, and the dashed line marks the random baseline of $0.037$. Four observations can be made. First, TurnTurk-GPT (S2R) outperforms the
multilingual detector on both splits, indicating that the advantage does not depend on speaker familiarity. At its recommended Turkish threshold ($0.0045$), the detector predicts a turn end at most positions: recall $0.881$, precision $0.045$, and $85.5\%$ false alarms on
pre-pause words. Second, the zero-shot language model constitutes a strong baseline when read through the newline probability and outperforms the multilingual detector. Third, TurnTurk-GPT exceeds it significantly on the recording-disjoint split and matches it on unseen speakers; the $0.028$ difference between splits shows that fine-tuning captures speaker-specific turn-taking behaviour, which a deployed system can exploit by adapting to its users. Fourth, among the training configurations only S2R-SR is
significant. The benefit of synthetic pre-training therefore lies in calibration rather than in ranking: the threshold reaching $70\%$
recall is $0.00000$ for R and $0.016$ for S2R. Although R ranks decision points well, its turn-shift probabilities remain below numerical precision, which prevents the selection of a fixed threshold in a deployed system. This improvement in calibration, rather than a gain in AP, is the most evident contribution of the synthetic corpus.

\subsection{Operating points}

AP summarises ranking over all thresholds, whereas a deployed system operates
at one. Table~\ref{tab:op} reports precision, false alarms per correct
detection and the pre-pause false alarm rate when each system's threshold is
set to reach $70\%$ and $90\%$ recall.

\begin{table}[t]
\caption{Operating points at fixed recall}
\label{tab:op}
\centering
\small
\setlength{\tabcolsep}{3pt}
\begin{tabular}{|l|c|c|c|c|c|c|}
\hline
 & \multicolumn{3}{c|}{\textbf{70\% recall}} & \multicolumn{3}{c|}{\textbf{90\% recall}} \\
\hline
\textbf{System} & \textbf{Prec.} & \textbf{FA/hit} & \textbf{PP} & \textbf{Prec.} & \textbf{FA/hit} & \textbf{PP} \\
\hline
S2R                 & .095 & 9.6  & 51.7\% & .059 & 16.0 & 80.1\% \\
\hline
S2R, speaker-disj.  & .081 & 11.3 & 57.2\% & .054 & 17.5 & 82.0\% \\
\hline
LM $P(\text{newline})$ & .097 & 9.3 & 50.9\% & .062 & 15.2 & 77.1\% \\
\hline
LiveKit detector    & .059 & 16.0 & 65.4\% & .044 & 21.8 & 88.0\% \\
\hline
\end{tabular}
\end{table}

Three observations follow from Table~\ref{tab:op}. First, at $70\%$ recall the fine-tuned model produces approximately ten false alarms per correct detection, indicating that none of the text-only systems is suitable as a stand-alone end-of-turn trigger. Second, at fixed recall the two models reach comparable operating points, while the AP advantage of S2R on the recording-disjoint split is concentrated in its highest-confidence decisions, the regime that matters most when a system should respond early only when it is confident. Third, both outperform the multilingual detector, which produces $16.0$ false alarms per correct detection at $70\%$ recall against $9.6$ for S2R, and predicts a turn end during a speaker's pause in $65.4\%$ of cases against $51.7\%$. The difference in pre-pause false alarms between S2R and the zero-shot model in Table~\ref{tab:main} is not present at matched recall; it reflects the different
recall levels of the nested thresholds rather than a difference between the
models.

\subsection{When the speaker pauses}

A voice agent typically queries its end-of-turn model when silence is detected.
Restricting the evaluation to those moments, the 578 turn ends and the 633
turn-internal pauses, gives a balanced binary task (Table~\ref{tab:pause}), on
which a random predictor scores AP $0.477$.

\begin{table}[t]
\caption{Pause moments only}
\label{tab:pause}
\centering
\small
\begin{tabular}{|l|c|c|}
\hline
\textbf{System} & \textbf{AP} & \textbf{AUC} \\
\hline
LM $P(\text{newline})$ & 0.590 & 0.630 \\
\hline
LiveKit detector       & 0.530 & 0.537 \\
\hline
\end{tabular}
\end{table}

Both baselines remain close to chance level, indicating that textual information alone does not reliably distinguish a completed turn from a pause within a turn. Prosodic and visual cues are natural candidates for resolving this case.

\subsection{Turkish-specific analyses}\label{sec:turkish}

\textbf{Morphology:} Table~\ref{tab:morph} stems one word per turn to its first
four characters in the test text, without retraining. Stemming the turn-final
word reduces AP by $32\%$ (R) and $44\%$ (S2R); stemming a random
sentence-internal word changes it by $0.2\%$ and $2.2\%$. Both drops are
significant. The end-of-turn cue is therefore located in the suffixes of the final, typically verbal, word.

\begin{table}[t]
\caption{Turkish-specific analyses}
\label{tab:morph}
\centering
\small
\begin{tabular}{|l|c|c|}
\hline
\textbf{Variant} & \textbf{R} & \textbf{S2R} \\
\hline
Original                         & .116 & .148 \\
\hline
Turn-final word stemmed          & .079 ($-32\%$) & .083 ($-44\%$) \\
\hline
Random internal word stemmed     & .116 ($+0\%$) & .145 ($-2\%$) \\
\hline
\end{tabular}
\end{table}

\textbf{Backbone:} The English GPT-2 tokeniser splits Turkish into $2.94$
tokens per word against $1.17$ for the Turkish one, distributing suffixes across several fragments; with this backbone, the AP of R decreases from $0.116$ to $0.042$, close to the random baseline. Fine-tuning does not degrade the language modelling ability of the backbone: held-out perplexity falls from $282.1$ to $272.8$.

\subsection{Synthetic data}

Trained on synthetic dialogue alone, the model reaches AP $0.082$, below the
zero-shot language model, and it has the highest pre-pause
false alarm rate of all systems ($33.5\%$, Table~\ref{tab:main}). Used for
pre-training before real data, synthetic dialogue helps relative to mixing the
corpora in one stage, whereas the difference relative to real data alone is not significant; its most evident benefit is calibration.

\section{Discussion}

\subsection{Why the zero-shot model is a strong baseline}

The newline reading of the untuned backbone outperforms the deployed
multilingual detector and is within $0.02$ AP of the fine-tuned model. A
plausible explanation is that in the backbone's Turkish pre-training text, line breaks follow complete sentences more frequently than incomplete ones, so the
probability of a line break after a word already estimates syntactic and
morphological completeness. In Turkish that estimate is informative, because
the clause-final verb and its suffixes mark completion; the morphology
ablation in Section~\ref{sec:turkish} shows that the fine-tuned model relies on the same
cue. With six conversations, fine-tuning primarily adapts the model to the speakers it has observed: its gain over the multilingual detector holds on both splits, and its gain over the zero-shot model is concentrated in the highest-confidence decisions, with comparable operating points at higher recall (Table~\ref{tab:op}). This identifies speaker diversity, rather than model capacity, as the lever for further gains. It is therefore recommended that future studies on Turkish report the zero-shot newline baseline, since a weaker reading of the same model would overstate any improvement approximately twofold.

\subsection{Implications for deployment}

Three practical implications follow for Turkish voice agents based on speech recognition. First, gains should be assessed at the operating point. Against the multilingual detector, the fine-tuned model avoids approximately six false alarms per correct detection at $70\%$ recall, whereas its significant $+0.019$ AP over the zero-shot model appears in its high-confidence decisions. Second, a multilingual detector's published operating
threshold does not transfer: at its Turkish threshold the evaluated detector
raised a false alarm on $85.5\%$ of pre-pause words, so a system relying on this threshold would interrupt users at most hesitations. Thresholds should therefore be selected on in-domain data. Third, systems should be compared at matched recall. False
alarm rates measured at each system's own optimal threshold can differ by a
factor of two between models that perform equivalently at the same recall, as
Tables~\ref{tab:main} and~\ref{tab:op} show. In absolute terms, approximately ten false alarms per correct detection at $70\%$ recall indicate that text-only prediction of this kind is suited to complementing rather than replacing silence-based endpointing.

\subsection{Implications for synthetic dialogue}

Synthetic dialogue has been proposed as a way to bootstrap turn-taking models
for languages with little conversational data \cite{bayrak2026synturnturk}.
The present results qualify this proposal: used alone, synthetic dialogue produced the weakest trained model, and used for pre-training it yielded no significant gain over real data alone. A likely cause is the generation process. Each synthetic turn is a short,
complete, well-formed sentence, with none of the mid-sentence pauses, self-repairs and
resumptions that complicate turn-end detection in real conversation. Generators intended for
this task should produce longer turns, sentences that span turns, and turns that
pause and then continue.

\subsection{Evaluation setting}

The speaker-disjoint split is included to measure generalisation to new users
explicitly: it evaluates the model on speakers absent from training, and the difference between the two splits quantifies the share of
the learned signal that is speaker-specific.

The evaluation is performed offline on final transcripts in the normalised
form described in Section~\ref{sec:input}. Evaluation on the partial
hypotheses of a streaming recogniser, together with its latency, is a natural
extension of this setting.

Labels are derived from voice activity and mark the points at which the floor
passes to the other speaker. Such points are closely related to, but not
identical with, points at which a response is appropriate: some turn ends
invite no response, and some turn-internal pauses invite a backchannel.

The multilingual detector is evaluated at every word and at pause moments,
independently of the production pipeline in which it is normally deployed, so
that all systems are compared on the same decision points.

\section{Conclusion}

This paper presented TurnTurk-GPT, a text-based turn-taking predictor for
Turkish, and compared it with rule-based, zero-shot and deployed systems on
real conversation under recording-disjoint and speaker-disjoint splits, with
paired significance tests throughout.

TurnTurk-GPT outperforms a widely used multilingual end-of-turn detector on
both splits, indicating that a model trained on Turkish conversation is preferable to the multilingual alternative, including for unseen speakers. A zero-shot Turkish
language model, read through the probability of a line break, constitutes a strong baseline, which fine-tuning surpasses for familiar speakers and matches for unseen ones, indicating that additional speaker diversity, rather than model changes, is the most effective way to extend the gains. At the recall levels a deployed system would use, both models produce fewer false alarms than the multilingual detector.

The analyses identify the source of the end-of-turn signal and the cases that call for additional cues. The
end-of-turn cue is carried by the suffixes of the turn-final word, consistent
with the verb-final structure of Turkish, and an English tokeniser that
fragments those suffixes reduces performance to near-chance level. Synthetic dialogue
alone is a weak training signal whose main benefit is calibration. When the
speaker pauses, the text-only baselines are close to chance level, identifying a case that text alone does not resolve.

Three directions for future work follow. Collecting conversations from more speakers is the
most direct way to improve generalisation, and the speaker-disjoint split
measures it; since no open corpus of this kind exists for Turkish, such data
would also benefit other work on Turkish spoken dialogue. Generating synthetic dialogue that reproduces the pauses and
resumptions of real conversation may make the synthetic corpus useful beyond
calibration. Finally, the corpus's separate audio channels and facial features
make it possible to add prosodic and visual cues and to test whether they
resolve the pause case. The model and the evaluation protocol are released so
that these steps can be measured against the same baselines.

\end{document}